\documentclass[runningheads]{llncs}
\usepackage[T1]{fontenc}
\usepackage{graphicx}
\usepackage{amsmath,amssymb}
\usepackage{algorithm}
\usepackage{algpseudocode}
\usepackage{booktabs}
\usepackage{makecell}
\usepackage{xcolor}
\usepackage{bm}
\usepackage{hyperref}
\usepackage{caption}


\begin{document}

\title{SDHN: Skewness-Driven Hypergraph Networks for Enhanced Localized Multi-Robot Coordination}
\titlerunning{SDHN: Skewness-Driven Hypergraph Networks}

\author{
Delin Zhao\inst{1} \and
Yanbo Shan\inst{1} \and
Chang Liu\inst{1} \and
Shenghang Lin\inst{1} \and
Yingxin Shou\inst{2}\thanks{Corresponding Author.} \and
Bin Xu\inst{1}
}
\authorrunning{D. Zhao et al.}

\institute{
School of Automation, Northwestern Polytechnical University, Xi'an, 710072, China\\
\email{delin\_zhao@qq.com} \and
School of Automation, Southeast University, Nanjing, 210096, China\\
\email{xueyediemeng@163.com}
}

\maketitle

\begin{abstract}
Multi-Agent Reinforcement Learning is widely used for multi-robot coordination, where simple graphs typically model pairwise interactions. However, such representations fail to capture higher-order collaborations, limiting effectiveness in complex tasks. While hypergraph-based approaches enhance cooperation, existing methods often generate arbitrary hypergraph structures and lack adaptability to environmental uncertainties. To address these challenges, we propose the Skewness-Driven Hypergraph Network (SDHN), which employs stochastic Bernoulli hyperedges to explicitly model higher-order multi-robot interactions. By introducing a skewness loss, SDHN promotes an efficient structure with Small-Hyperedge Dominant Hypergraph, allowing robots to prioritize localized synchronization while still adhering to the overall information, similar to human coordination. Extensive experiments on Moving Agents in Formation and Robotic Warehouse tasks validate SDHN’s effectiveness, demonstrating superior performance over state-of-the-art baselines.
\keywords{Multi-Agent Reinforcement Learning \and Hypergraph Neural Network \and Multi-Robot Coordination \and Skewness Loss}
\end{abstract}

\section{Introduction}
 
Achieving effective collaboration in multi-robot systems remains a significant challenge in complex environments, where traditional coordination methods struggle with scalability and generalization~\cite{rw1,rw3}. In recent years, Multi-Agent Reinforcement Learning (MARL) has emerged as a powerful paradigm for addressing multi-robot coordination problems~\cite{rw7}, with notable frameworks including Centralized Training with Decentralized Execution (CTDE) approaches such as MADDPG~\cite{rw9} and MAPPO~\cite{rw10}, and value decomposition methods like QMIX~\cite{rw11} and QPLEX~\cite{rw12}. These methods offer advantages in distributed decision-making and complex interaction modeling~\cite{it1}, but they still face limitations in capturing intricate coordination patterns.
 
In multi-robot tasks, effective collaboration requires a structured representation of cooperative relationships among robots. A common approach is to model these relationships using graph-based representations, where robots are represented as nodes and their interactions as edges~\cite{it4}. However, existing graph-based methods such as DCG~\cite{rw14}, DICG~\cite{rw15}, and CASEC~\cite{rw16} predominantly rely on simple graphs that capture only pairwise interactions. Consequently, these approaches are limited in accounting for higher-order coordination patterns, which are crucial for multi-robot applications such as multi-UAV cooperative pursuit~\cite{it2} and multi-robot formation control~\cite{it3}. These scenarios require interactions involving three or more robots simultaneously, yet simple graph structures inherently lack the capacity to model such complex dependencies due to their pairwise-edge constraints.
 
To overcome this limitation, recent studies such as HyperComm~\cite{rw24}, HGAC~\cite{rw25}, and SAEIR~\cite{rw26} have introduced hypergraphs, where hyperedges can connect multiple agents and explicitly model higher-order collaboration. Despite their potential, existing hypergraph-based approaches face two critical challenges: \textbf{(1) Lack of Informative Hyperedge Construction}: Current methods often generate hyperedges arbitrarily, without leveraging multi-robot collaborative priors. This results in hypergraph topologies that may fail to capture essential coordination patterns, ultimately limiting effectiveness. \textbf{(2) Deterministic Hyperedge Representation}: Existing approaches typically use deterministic hypergraphs to model collaboration. However, multi-robot systems operating in dynamic and stochastic environments are inherently noisy and uncertain. Rigid hypergraph structures struggle to adapt to such variability, reducing algorithmic robustness.
 
\begin{figure}[t]
\centering
\includegraphics[width=0.6\columnwidth]{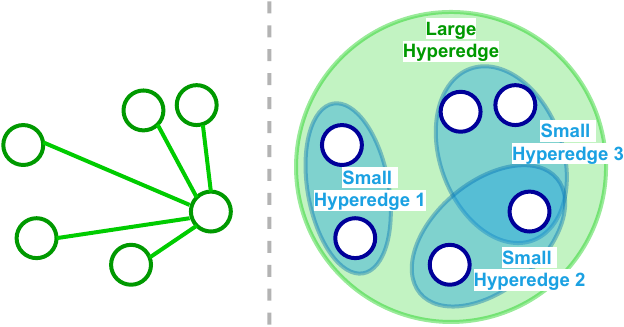} 
\caption{Circular nodes represent robots. \textbf{Left}: Each pairwise edge (line) connects two robots. \textbf{Right}: Each hyperedge (circle or ellipse) connects multiple robots. SDHN promotes a hypergraph structure with \textit{fewer large hyperedges and more small hyperedges} for efficient interaction, the structure we term the Small-Hyperedge Dominant Hypergraph.}
\label{fig:Motivation}
\vspace{-15pt} 
\end{figure}
 
To address these limitations, this paper proposes SDHN, a multi-robot decision-making framework that integrates hypergraph theory with MARL. SDHN models both individual and collective collaboration by introducing a skewness-driven coordination mechanism, which prioritizes small-team interactions while maintaining global cooperation. For example, in multi-UAV formation flight, SDHN enables drones to prioritize close-range synchronization while still adhering to the overall formation pattern, allowing for adaptive local adjustments without disrupting global stability. Additionally, SDHN represents hyperedges as Bernoulli distributions to enhance adaptability to environmental uncertainties. The framework follows a CTDE paradigm: during training, a centralized critic first generates stochastic hypergraphs via distribution sampling, and then a Hypergraph Convolutional Network (HGCN) performs message passing on these hypergraphs. During execution, individual agents make decisions independently based on their local observations.
 
The main contributions of this work are summarized as follows:
\begin{itemize}
    \item Skewness-Driven Hypergraph Generation: We introduce a Small-Hyperedge Dominant Hypergraph, a structured hypergraph that incorporates topological constraints to enhance coordination efficiency.
    \item Stochastic Hyperedge Representation: We develop a probabilistic hypergraph approach that improves robustness and expressiveness, particularly in environments with uncertainty.
    \item Extensive Empirical Validation: Experiments on Moving Agents in Formation (MAIF) and Robotic-Warehouse (RWARE) tasks demonstrate SDHN’s superior performance, highlighting its effectiveness in multi-robot coordination.
\end{itemize}

\section{Preliminary}

\subsection{Background on MARL}

The coordination of a multi-robot system with $n$ homogeneous agents can be modeled as a Decentralized Partially Observable Markov Decision Process (Dec-POMDP) defined by the tuple 
$\langle \mathcal{A}, \mathcal{S}, \{\mathcal{U}_i\}_{i=1}^n, P, \{\mathcal{O}_i\}_{i=1}^n, \{\pi_i\}_{i=1}^n, R, \gamma \rangle,$
where $\mathcal{A}$ denotes the set of $n$ identical agents,  $s \in \mathcal{S}$ represents the true environmental state, and $\gamma \in [0,1)$ is the discount factor. Each agent $a_i \in \mathcal{A}$ partially observes the environment through local observations $o_i^t \in \mathcal{O}_i$ and executes actions $u_i^t \in \mathcal{U}_i$ according to their individual policies $\pi_i(u_i^t|o_i^t)$. The joint action $\mathbf{u} = (u_1, \dots, u_n)$ induces state transitions governed by the dynamics $P(s'|s,\mathbf{u})$. A shared reward signal $R(s,\mathbf{u})$ evaluates the collective performance. The objective is to learn policies $\{\pi_i\}_{i=1}^n$ that collectively maximize the expected discounted return
\[
E_{s_{0:\infty}, \mathbf{u}_{0:\infty}} \sum_{t=0}^\infty \gamma^t R(s^t, \mathbf{u}^t) 
\]

MAPPO extends the Proximal Policy Optimization framework~\cite{ppo} by incorporating a centralized value function that aggregates global information, improving stability and efficiency in cooperative MARL tasks~\cite{rw10}. The policy network $\bm{\theta}$ is optimized by maximizing the following objective:
\begin{equation}
    J(\bm{\theta}) = \sum_{i=1}^N \mathbb{E} \left[ \min \left( \eta^i_t(\bm{\theta})\, \hat{A}^i_t,\, \operatorname{clip}\left( \eta^i_t(\bm{\theta}),\, 1\pm\epsilon \right)\, \hat{A}^i_t \right) \right]
    \label{eq:mappo}
\end{equation}
where $\eta^i_t(\bm{\theta}) = \frac{\pi_{\bm{\theta}}(a^i_t \mid \tau^i_t)}{\pi_{\bm{\theta}_{\text{old}}}(a^i_t \mid \tau^i_t)}$ is the probability ratio. $\text{clip}(\cdot)$ restricts $\eta^i_t(\theta)$ to the interval $[1-\epsilon, 1+\epsilon]$. $\hat{A}^i_t$ is the Generalized Advantage Estimation (GAE)~\cite{gae}. $\hat{A}^i_t = \sum_{l=0}^T (\gamma \lambda)^l A^i_{t+l}$, where $T$ is the trajectory horizon, $\gamma$ is the discount factor, and $\lambda$ is the GAE hyperparameter. Although MAPPO improves multi-agent coordination, it primarily focuses on decentralized decision-making with shared global information, which can be limiting in tasks that require higher-order interactions.

\begin{figure}[t]
\centering
\includegraphics[width=\textwidth]{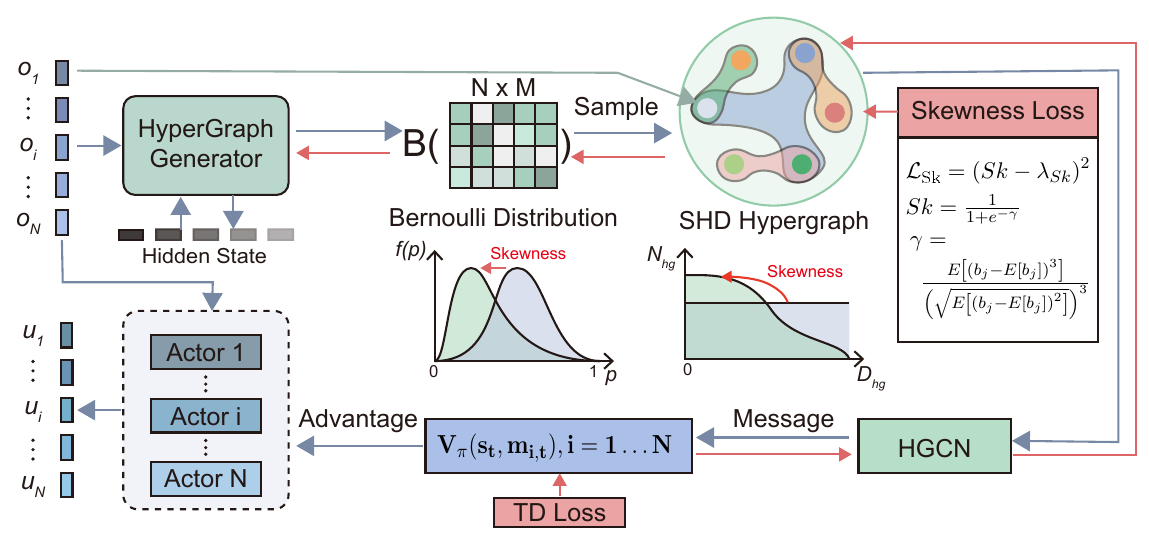}
\caption{The framework of proposed method. The $f(p)$--$p$ curve displays the random parameter $p$ generated by the hypergraph generator along with its probability density $f(p)$, where the skewness constraint biases the distribution toward smaller values. Meanwhile, the $N_{\text{hg}}$--$D_{\text{hg}}$ curve indicates that $D_{\text{hg}}$ (the hyperedge degree) and $N_{\text{hg}}$ (the hyperedge count) are guided by the skewness mechanism to yield more small hyperedges than large ones. SHD Hypergraph represents the Small-Hyperedge Dominant Hypergraph.}
\label{fig:framework}
\vspace{-15pt} 
\end{figure}

\subsection{Hypergraph Representation}
A hypergraph generalizes simple graphs by allowing edges to connect multiple vertices, enabling the representation of group-based interactions. Mathematically, a hypergraph is defined as $G = (\mathcal{V}, \mathcal{E})$,
where \(\mathcal{V} = \{v_1, \ldots, v_N\}\) denotes the set of vertices with \(N\) representing the number of agents, and \(\mathcal{E} = \{e_i = (v_{1}^{i}, v_{2}^{i}, \dots, v_{N}^{i}) \mid i=1,2,\dots, M\}\)
denotes the set of hyperedges, with \(M\) being the number of hyperedges. In the context of multi-robot systems, each vertex corresponds to an individual robot, and a hyperedge \(e \in \mathcal{E}\) models a higher-order interaction among a subset of agents. This structure allows for the capture of complex group behaviours that are critical in tasks like formation control~\cite{rw19}, real-time strategy~\cite{rw20}, and collective search~\cite{rw21}.

\section{Method}

SDHN adopts an Actor-Critic-based CTDE architecture, where the centralized critic utilizes global state information during training while the actor makes decisions based on local observations during execution. The critic fuses current observations with a hidden state enriched by historical data, allowing it to capture cooperative relationships at multiple levels. These relationships are then represented as a hypergraph, where each hyperedge is modelled by a stochastic distribution to account for the inherent uncertainty and variability in the interactions. To facilitate decision-making, a hypergraph convolutional network (HGCN) enables the exchange of information among agents. Additionally, a skewness loss function guides the hypergraph generator to form a more efficient cooperative structure by prioritizing smaller, more localized hyperedges. The overall framework of our proposed approach is shown in Fig.~\ref{fig:framework}.

\subsection{Hypergraph Generation and Inference Mechanism}

A hypergraph generator \(f_{g}(\cdot)\) is designed, which extracts a latent state of the group from the agents' observations and transforms it into a hyperedge distribution matrix \(\mathbf{P_H}\):
\begin{equation}
\mathbf{P_H} = f_g\left(\{\mathbf{o}_i^t\}_{i=1}^N, \mathbf{h}^{t-1}\right)
\label{eq:Hypergraph_generator}
\end{equation}
where \(\{\mathbf{o}_i^t\}_{i=1}^N\) denotes the observations of the agents. To capture the sequential nature of these observations, \(f_{g}(\cdot)\) is implemented using Recurrent Neural Networks (RNN), with \(\mathbf{h}^{t-1}\) serving as the historical hidden state. The matrix \(\mathbf{P_H} \in [0,1]^{N \times M}\) contains the parameters of Bernoulli distributions that represent each hyperedge, where \(M\) is the number of hyperedges.

After obtaining $\mathbf{P_H}$, a hyperedge incidence matrix \(\mathbf{H} \in \{0,1\}^{N \times M}\) (with \(\mathbf{H}_{v,e}=1\) if vertex \(v\) belongs to hyperedge \(e\), and \(\mathbf{H}_{v,e}=0\) otherwise) can be sampled. To enable differentiable sampling, we adopt the Binary Concrete relaxation with Gumbel reparameterization. We first compute Gumbel noise $\epsilon = \log(-\log(u))$, where \(u \sim \mathcal{U}(0,1)\). Applying the sigmoid function yields a differentiable sample:
\begin{equation}
    y =\sigma\left(\frac{\log(p)-\log(1-p)+\epsilon}{\tau}\right)
\end{equation}
where \(\sigma(\cdot)\) is the sigmoid function and \(\tau\) is the temperature parameter.
The discrete hyperedge assignment \(H(v,e) = \mathbb{I}\{y \ge 0.5\}\) constructs the incidence matrix \(\mathbf{H}\), ensuring end-to-end differentiability for generator training. 

Subsequently, an HGCN is employed to propagate and aggregate information across the hypergraph, thereby facilitating the effective sharing and integration of information among agents and augmenting cooperative strategies. The HGCN operation is defined as:
\begin{equation}
\mathbf{X}^{(l+1)} = \operatorname{ReLU}\left(\mathbf{D}^{-1/2} \mathbf{H} \mathbf{B}^{-1} \mathbf{H}^\mathrm{T} \mathbf{D}^{-1/2} \mathbf{X}^{(l)} \mathbf{\Theta}\right)
\label{eq:HGCN}
\end{equation}
where \(\mathbf{X}^{(l)} \in \mathbb{R}^{N \times F^{(l)}}\) and \(\mathbf{X}^{(l+1)} \in \mathbb{R}^{N \times F^{(l+1)}}\) denote the inputs of the \(l\)-th and \((l+1)\)-th layers, respectively, with \(\mathbf{X}^{(l+1)}\) being the extracted observation features \(\{\hat{o}_1^t, \ldots, \hat{o}_N^t\}\). The weight matrix between the two layers is \(\mathbf{\Theta} \in \mathbb{R}^{F^{(l)} \times F^{(l+1)}}\). The degree matrices $\mathbf{D}\in\mathbb{R}^{N\times N}$ and $\mathbf{B}\in\mathbb{R}^{M\times M}$ for the vertices and hyperedges are defined as $\mathbf{D}=\operatorname{diag}(d_1,\dots,d_N)$ with $d_i=\sum_{j=1}^M \mathbf{H}_{ij}$ for $i=1,\dots,N$; and $\mathbf{B}=\operatorname{diag}(b_1,\dots,b_M)$ with $b_j=\sum_{i=1}^N \mathbf{H}_{ij}$ for $j=1,\dots,M$.

The output of the HGCN, denoted as \(\mathbf{m}^{t} = \mathbf{X}_l^t\), represents the aggregated features from collaborative agents. Specifically, \(\mathbf{m}^{t}_i\) denotes the cooperative feature obtained by agent \(i\), which includes information about the interactions and dependencies with other agents in the system. When \(\mathbf{m}^{t}_i\) is input into the Q-value function, it enables agent \(i\) to make more informed decisions. This facilitates better coordination, as each agent learns to align its actions with the collective objectives of the group, improving overall system performance in complex multi-agent environments.

\subsection{Skewness-Driven Hypergraph MARL}

Inspired by natural group behaviors(e.g., group hunting, migration), we propose a skewness-driven hypergraph generator that yields Small-Hyperedge Dominant Hypergraphs: agents interact intensely with a few neighbors while retaining global awareness.

Hyperedge skewness is defined as:
\begin{equation}
    Sk = \frac{2}{1 + e^{-\gamma}} - 1
\end{equation}
\begin{equation}
    \gamma = \frac{\mathbb{E}\left[\left(b_{j} - \mathbb{E}[b_{j}]\right)^3\right]}{\left(\sqrt{\mathbb{E}\left[\left(b_{j} - \mathbb{E}[b_{j}]\right)^2\right]}\right)^3}
\end{equation}
where \(b_j\) is the degree of hyperedge \(j\) and \(Sk \in (-1,1)\) indicates the bias in hyperedge sizes. A value of \(Sk\) near \(-1\) indicates a predominance of small hyperedges, whereas \(Sk\) approaching 1 larger hyperedges are more dominant.

Based on this, skewness loss is constructed as:
\begin{equation} 
\mathcal{L}_{\text{Sk}} = \left( Sk - \lambda_{Sk} \right)^2 \label{eq:skew_loss} 
\end{equation}
where $\lambda_{Sk}$ is a target parameter for the skewness. This loss function adjusts the third-order moment of hyperedge connectivity, encouraging the network to prioritize interactions among a small subset of agents while preserving awareness of the team’s state.  

In the proposed algorithm, the critic estimates the state value based on the global state, while the actor is trained using the same approach as in MAPPO. The critic is jointly trained using a temporal-difference (TD) loss and skewness loss:
\begin{equation}
\mathcal{L}_{\text{total}}(\boldsymbol{\theta}) = \mathcal{L}_{\text{TD}}(\boldsymbol{\theta}) + \lambda_{CB} \, \mathcal{L}_{\text{Sk}}(\boldsymbol{\theta}_{hg})    \label{eq:total_loss}
\end{equation}
where \(\boldsymbol{\theta}\) includes all the parameters in the critic network, \(\boldsymbol{\theta}_{hg}\) denotes the subset of the parameters specifically related to hypergraph. \(\lambda_{CB}\) is the character-balance weight assigned to the skewness loss. The TD loss \(\mathcal{L}_{\text{TD}}(\boldsymbol{\theta})\) is defined as:
\begin{equation}
\mathcal{L}_{\text{TD}}(\boldsymbol{\theta}) = \mathbb{E}_{s^t, \mathbf{u}^t}\left[\left(V(s^t;\boldsymbol{\theta}) - V_t^{\text{'}}\right)^2\right]
\label{eq:loss}
\end{equation}
\begin{equation}
V_t^{\text{'}} = A_t^{\text{GAE}(\gamma,\lambda)} + V(s^t;\boldsymbol{\theta}')
\label{eq:v_target}
\end{equation}
The GAE advantage \(A_t^{\text{GAE}(\gamma,\lambda)}\) is defined as:
\begin{equation}
A_t^{\text{GAE}(\gamma,\lambda)} = \sum_{l=0}^{\infty} (\gamma\lambda)^l \delta^{t+l}
\label{eq:gae_advantage}
\end{equation}
\begin{equation}
\delta^t = r^t + \gamma V(s^{t+1}; \boldsymbol{\theta}') - V(s^t; \boldsymbol{\theta}')
\label{eq:td_error}
\end{equation}
where \(\gamma\) is the discount factor and \(\lambda\) is the weight factor for GAE. \(\boldsymbol{\theta}'\) represents a target copy of the critic parameters used for stabilizing the TD target computation.


\section{Experiments}

To evaluate the effectiveness of the proposed SDHN algorithm, we designed a series of experiments to address the following questions: (1) How does SDHN compare to SOTA methods across tasks? (Section~\ref{sec:Baselines_Compare}) (2) How does SDHN's performance scale with different numbers of robots? (Section~\ref{sec:Scalability_Analysis}) (3) What is the impact of the skewness loss? (Section~\ref{sec:Ablation_Skewness}) (4) Does the stochastic hyperedge representation help? (Section~\ref{sec:Ablation_SHR})

Evaluated on MAIF~\cite{rw19} and RWARE~\cite{RWARE}, SDHN is benchmarked against SOTA prior methods (Table~\ref{tab:method_comparison}) under varying robot counts and scenarios. Two ablations test the skewness loss ($\mathcal{L}_{\text{Sk}}$) and the stochastic hyperedge representation (SHR). 

We employ distinct 3-layer RNN in Eq.~(\ref{eq:Hypergraph_generator}) and 2-layer HGCN as specified in Eq.~(\ref{eq:HGCN}). The target parameter $\lambda_{Sk}$ in Eq.~(\ref{eq:skew_loss}) is set to -0.6 and the character-balance weight \(\lambda_{CB}\) in Eq.~(\ref{eq:total_loss}) is set to 1. \textcolor{black}{More experimental details and our implementation can be found at \textit{https://github.com/DeLin1001/SDHN}}.




\begin{figure}[t]
\centering
\includegraphics[width=\textwidth]{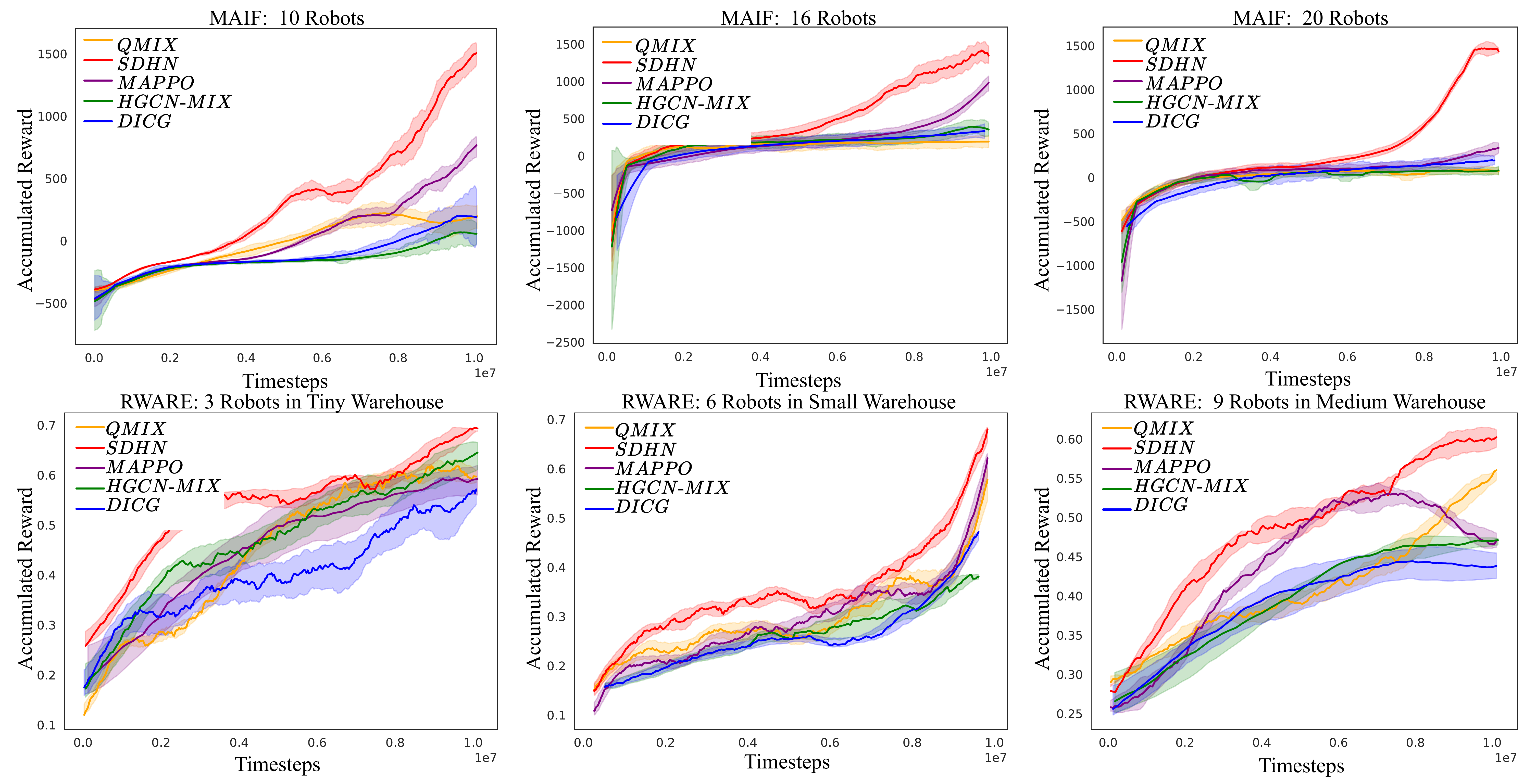}
\caption{Performance of SDHN and baselines on six environments from two tasks.}
\label{fig:RESULT_SDHN}
\vspace{-5pt} 
\end{figure}

\subsection{Benchmarking and Scalability Evaluation}

\begin{figure}[t]
\centering
\begin{minipage}{0.48\columnwidth}
  \centering
  \includegraphics[width=0.8\linewidth]{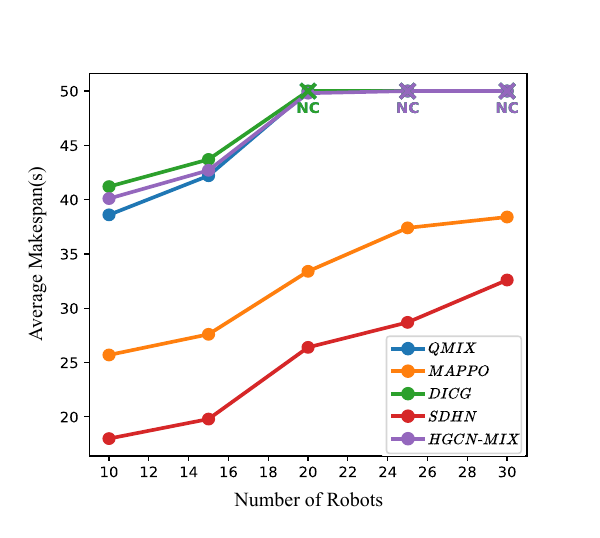}
  \caption{Average makespan comparison on MAIF tasks. NC=Not Completed.}
  \label{fig:Makespan}
\end{minipage}%
\hfill
\begin{minipage}{0.48\columnwidth}
  \centering
  \captionof{table}{Method comparison on graph modeling. Det.=deterministic, Stoch.=stochastic, Rep.=represent.}
  \label{tab:method_comparison}
  \begin{tabular}{lccc}
    \toprule
    Method & Graph & Edge & Prior \\[-2pt]  
           & Type  & Rep. & Guide\\
    \midrule
    QMIX~\cite{rw11}     & -- & -- & --\\
    MAPPO~\cite{rw10}    & -- & -- & --\\
    DICG~\cite{rw15}     & Simple & Det. & --\\
    HGCN-MIX~\cite{rw27} & Hyper & Det. & --\\
    SDHN                 & Hyper & Stoch. & Yes\\
    \bottomrule
  \end{tabular}
\end{minipage}
\vspace{-15pt} 
\end{figure}

\textbf{Comparison with Baselines} \label{sec:Baselines_Compare}
In Fig.~\ref{fig:RESULT_SDHN}, we present a comprehensive overview of our experiments conducted across six environments, highlighting the superior performance of SDHN. SDHN consistently achieves high accumulated rewards with rapid convergence and stability, outperforming competing methods across a variety of tasks and scenarios of differing complexity—thus demonstrating its generalizability. In the \textit{RWARE: 3 Robots} and \textit{RWARE: 6 Robots} environments, all methods exhibit similar convergence patterns. However, in other settings the differences become pronounced: in the \textit{MAIF: 10 Robots} and \textit{RWARE: 9 Robots} environments, HGCN-MIX and DICG struggle to match the performance of the leading methods, while in the \textit{MAIF: 16 Robots} and \textit{MAIF: 20 Robots} environments, QMIX and MAPPO fail to produce high-reward strategies. 
\\ \\
\textbf{Scalability Evaluation} \label{sec:Scalability_Analysis}
Fig.~\ref{fig:Makespan} presents the average makespan under progressively increasing robot counts for both SDHN and the baseline methods. Although the makespan naturally rises with a higher number of robots across all methods, SDHN consistently achieves the shortest makespan, thereby demonstrating its superior efficiency and scalability. Notably, SDHN achieved a 23\% reduction in makespan compared to MAPPO---the best-performing baseline among the compared methods. By contrast, while QMIX, DICG, and HGCN-MIX exhibit moderate improvements in reward accumulation during training, they fail to learn effective formation strategies, resulting in considerably longer task completion times and, in many cases, an inability to complete the MAIF task as the robot population increases.

The limitations of the compared methods are evident: QMIX and MAPPO lack graph structures, which limits their ability to model collaborative relationships; although DICG employs a simple graph to capture these relations, it falls short of representing higher-order interactions. HGCN-MIX leverages a hypergraph to represent collaboration but lacks prior information, resulting in an arbitrary hypergraph structure that may fail to capture effective interactions. Furthermore, both DICG and HGCN-MIX rely on deterministic graph structures, which further restrict their capacity to express collaborative relationships, leading to suboptimal performance. In contrast, the advantage of SDHN over MAPPO underscores the effectiveness of our hypergraph-based approach for modeling collaboration. Table~\ref{tab:method_comparison} provides a detailed overview of the design characteristics across the compared methods.

In summary, comparative experiments demonstrate that SDHN delivers exceptional performance in MARL. Moreover, scalability evaluations reveal that SDHN maintains robust performance as the number of robots increases, achieving notably reduced makespans relative to other baselines. By leveraging the skewness-driven hypergraph generation and stochastic hyperedge representation, SDHN offers a smarter and more adaptable solution for multi-robot collaboration.

\subsection{Ablation Study}

\begin{figure}[t]
\centering
\includegraphics[width=0.8\columnwidth]{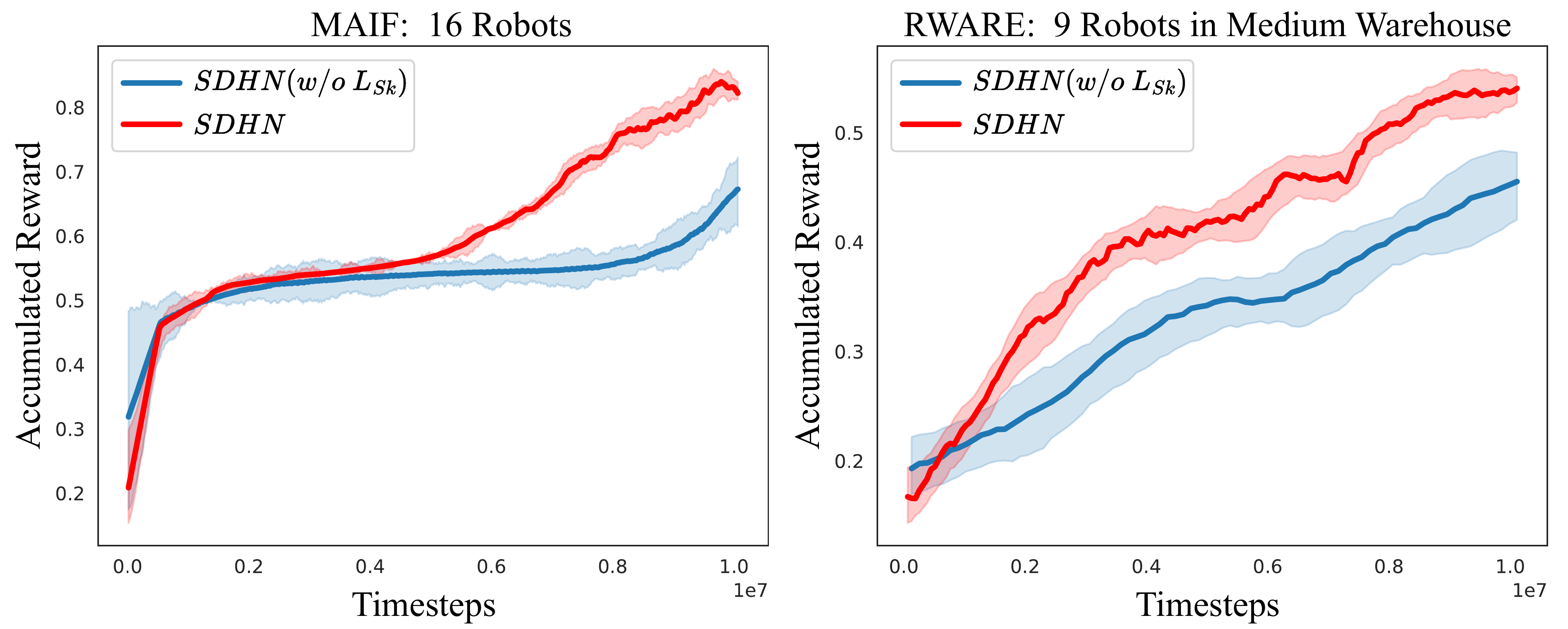}
\caption{Experiments comparing training with and without the skewness loss. }
\label{fig:Ablation_WOLSK}
\vspace{-15pt} 
\end{figure}

We further investigate the contributions of our key components through comprehensive ablation studies on multiple maps, focusing on two aspects: the effectiveness of the skewness loss and that of the stochastic hyperedge representation.

\begin{figure}[t]
\centering
\includegraphics[width=0.8\columnwidth]{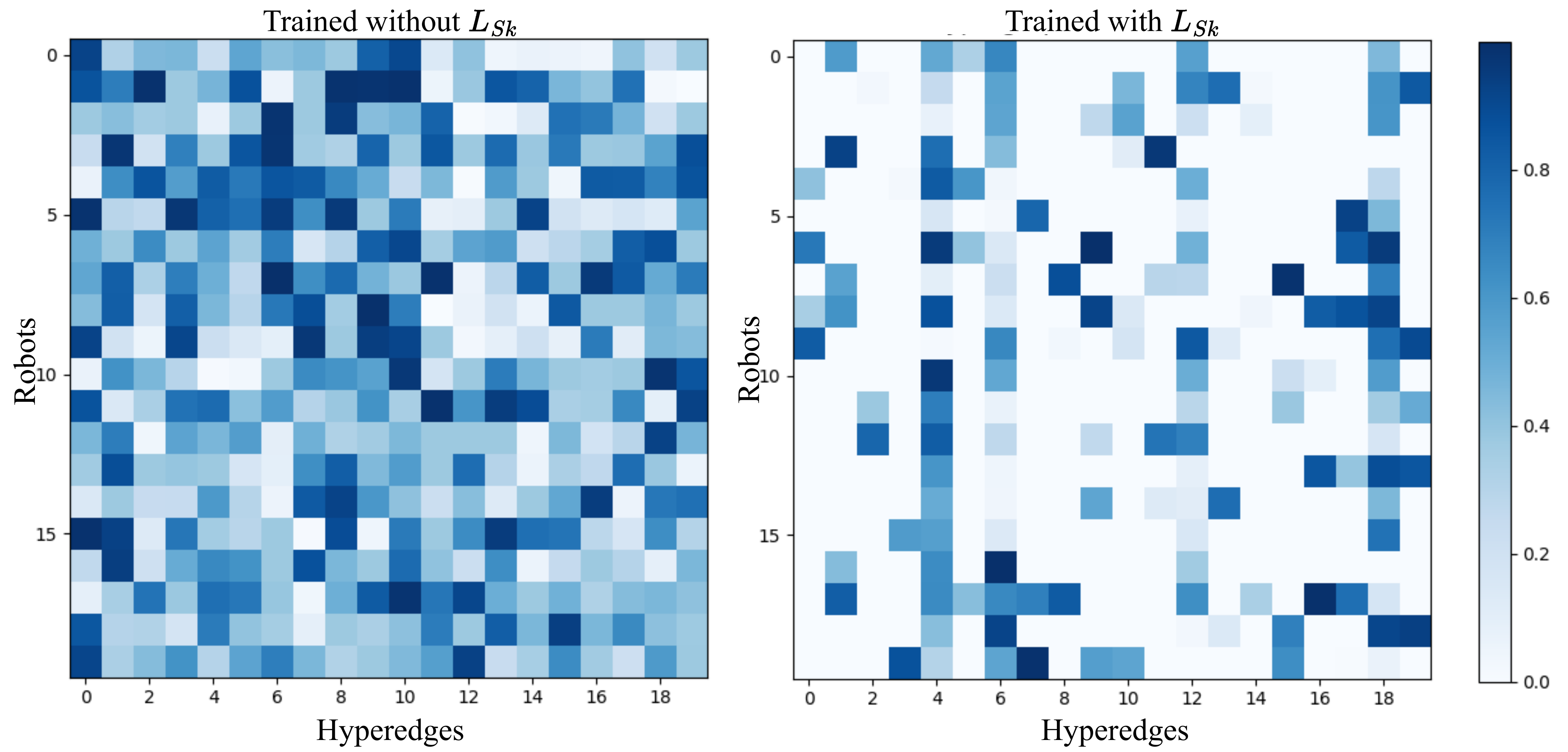}
\caption{Bernoulli hyperedge distribution matrix $P_{H}$ for models trained without and with the skewness loss $\mathcal{L}_{Sk}$. \textbf{Left}: The hyperedge probability matrix obtained without $\mathcal{L}_{Sk}$ exhibits a more even, random distribution; \textbf{Right}: After training with the skewness loss, low probabilities become dominant, indicating a learned preference for generating a Small-Hyperedge Dominant Hypergraph.}
\label{fig:PH}
\vspace{-10pt} 
\end{figure}

Effectiveness of Skewness Loss\label{sec:Ablation_Skewness}: As shown in Fig.~\ref{fig:Ablation_WOLSK}, incorporating the skewness loss $\mathcal{L}_{Sk}$ markedly improves the performance of SDHN. Specifically, SDHN trained with $\mathcal{L}_{Sk}$ not only converges faster with lower variance but also achieves a higher final average performance compared to the variant without the skewness loss. These results indicate that the skewness loss effectively guides the training of the hypergraph generator. It promotes the formation of a Small-Hyperedge Dominant Hypergraph, which in turn fosters more effective cooperative learning among agents. Fig.~\ref{fig:PH} provides clear visual evidence of this mechanism, demonstrating that the skewness loss systematically induces a Small-Hyperedge Dominant Hypergraph.

\begin{figure}[t]
\centering
\includegraphics[width=0.8\columnwidth]{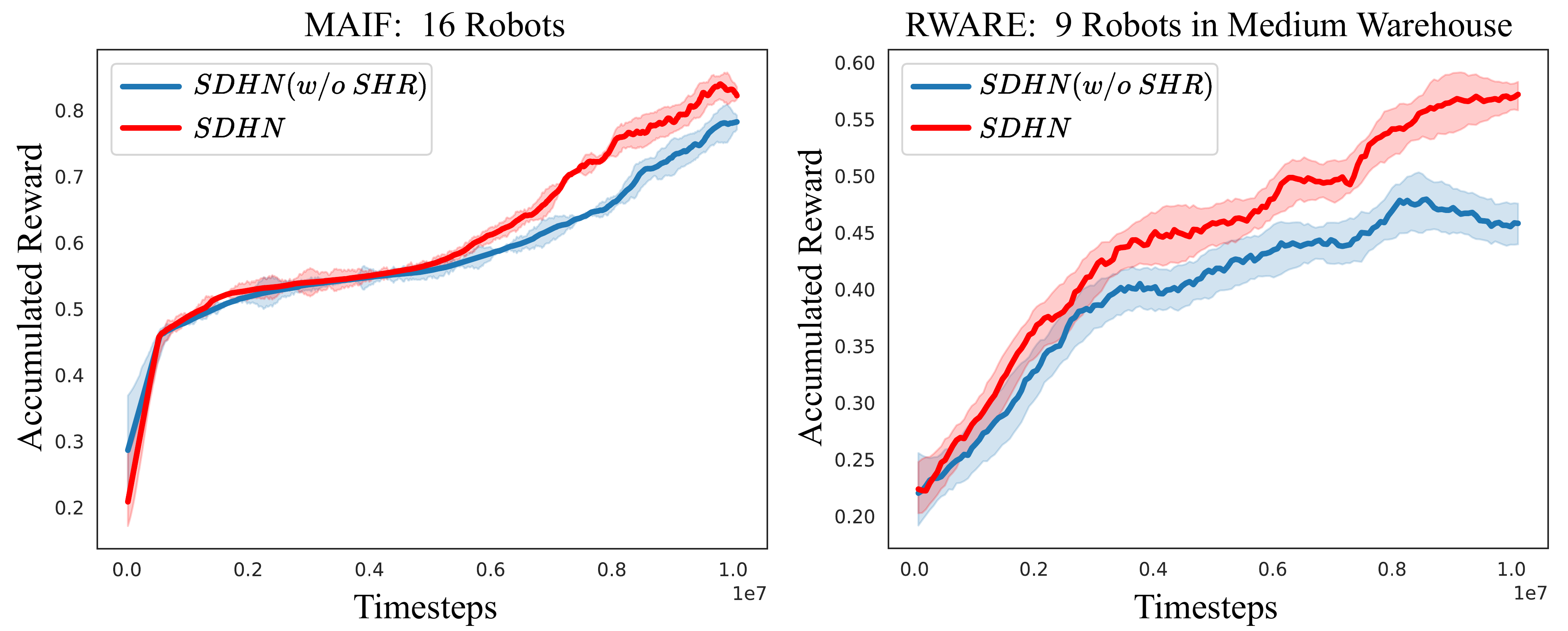}
\caption{Experiments comparing training SDHN with and without a stochastic hyperedge representation. }
\label{fig:Ablation_WOSHR}
\vspace{-15pt}
\end{figure}

Effectiveness of Stochastic Hyperedge Representation:\label{sec:Ablation_SHR}
We also evaluate the impact of employing a stochastic hyperedge representation on two different maps. As demonstrated in Fig.~\ref{fig:Ablation_WOSHR}, SDHN with a stochastic hyperedge representation consistently outperforms the deterministic counterpart. In the absence of a stochastic hyperedge representation, the model struggles to discover optimal strategies, particularly under complex cooperative scenarios. This finding suggests that modeling hyperedges as Bernoulli distributions enables the network to better capture environmental uncertainties, thereby enhancing the expressive power of collaborative relationships and improving overall performance.

\section{Conclusion}

In this paper, we propose SDHN, a novel approach for multi-robot coordination via MARL. By introducing a skewness loss, SDHN guides the formation of efficient cooperative structures, effectively captures both local and global collaboration patterns. This design allows robots to focus on the most relevant peers while maintaining awareness of the overall state, mimicking real-world coordination dynamics in tasks such as group hunting or migration. Moreover, SDHN utilizes stochastic Bernoulli hyperedges to model higher-order interactions, where the inherent randomness effectively models environmental uncertainties and enables dynamic adaptation of the collaboration structure. To validate our approach, we conducted extensive experiments on MAIF and RWARE tasks including benchmarking, scalability evaluations and ablation studies. These comprehensive evaluations show that SDHN outperforms current state-of-the-art methods, highlighting the value of guided hypergraph structures and stochastic hyperedge representations in multi-robot systems. Future work focuses on real-world deployment.

\bibliographystyle{splncs04}
\bibliography{references}

\begin{thebibliography}{10}
\providecommand{\url}[1]{\texttt{#1}}
\providecommand{\urlprefix}{URL }
\providecommand{\doi}[1]{https://doi.org/#1}

\bibitem{rw27}
Bai, Y., Gong, C., Zhang, B., Fan, G., Hou, X., Lu, Y.: Cooperative multi-agent reinforcement learning with hypergraph convolution. In: International Joint Conference on Neural Networks (IJCNN). pp.~1--8 (2022). \doi{10.1109/IJCNN55064.2022.9891942}

\bibitem{rw14}
Boehmer, W., Kurin, V., Whiteson, S.: Deep coordination graphs. In: Thirty-Seventh International Conference on Machine Learning (ICML). pp. 980--991 (2020)

\bibitem{it4}
Duan, W., Lu, J., Xuan, J.: Inferring latent temporal sparse coordination graph for multiagent reinforcement learning. IEEE Transactions on Neural Networks and Learning Systems p. 1–13 (2024). \doi{10.1109/tnnls.2024.3513402}, \url{http://dx.doi.org/10.1109/tnnls.2024.3513402}

\bibitem{rw20}
Ellis, B., Cook, J., Moalla, S., Samvelyan, M., Sun, M., Mahajan, A., Foerster, J.N., Whiteson, S.: Smacv2: An improved benchmark for cooperative multi-agent reinforcement learning. In: Advances in Neural Information Processing Systems (NeurIPS) (2023)

\bibitem{rw3}
Guo, P., Zhang, R., Xu, B.: Safety separation distance design for uav formation based on system performance. IEEE Transactions on Aerospace and Electronic Systems  \textbf{61}(3),  5980–5995 (Jun 2025). \doi{10.1109/taes.2024.3524948}, \url{http://dx.doi.org/10.1109/taes.2024.3524948}

\bibitem{rw26}
He, X., Ge, H., Hou, Y., Yu, J.: Saeir: Sequentially accumulated entropy intrinsic reward for cooperative multi-agent reinforcement learning with sparse reward. In: Thirty-Third International Joint Conference on Artificial Intelligence (IJCAI). pp. 4107--4115 (2024). \doi{10.24963/ijcai.2024/454}

\bibitem{rw7}
Huang, Z., Yang, Z., Krupani, R., Senbaslar, B., Batra, S., Sukhatme, G.S.: Collision avoidance and navigation for a quadrotor swarm using end-to-end deep reinforcement learning. In: IEEE International Conference on Robotics and Automation (ICRA). pp. 300--306 (2024). \doi{10.1109/ICRA57147.2024.10611499}

\bibitem{rw15}
Li, S., Gupta, J.K., Morales, P., Allen, R.E., Kochenderfer, M.J.: Deep implicit coordination graphs for multi-agent reinforcement learning. In: 20th International Conference on Autonomous Agents and Multiagent Systems (AAMAS). pp. 764--772 (2021). \doi{10.5555/3463952.3464044}

\bibitem{rw1}
Li, S., Batra, R., Brown, D., Chang, H.D., Ranganathan, N., Hoberman, C., Rus, D., Lipson, H.: Particle robotics based on statistical mechanics of loosely coupled components. Nature  \textbf{567}(7748),  361--365 (2019). \doi{10.3410/f.735358013.793558239}

\bibitem{it3}
Li, X., Zhou, R., Zhang, Y., Sun, G.: Distributed shape formation of multirobot systems via dynamic assignment. IEEE Transactions on Industrial Electronics  \textbf{72}(3),  3017--3027 (2025). \doi{10.1109/tie.2024.3436657/mm1}

\bibitem{rw19}
Lin, Q., Ma, H.: Mfc-eq: Mean-field control with envelope q-learning for moving decentralized agents in formation. In: IEEE/RSJ International Conference on Intelligent Robots and Systems (IROS). pp. 14156--14163 (2024). \doi{10.1109/IROS58592.2024.10802293}

\bibitem{rw9}
Lowe, R., Wu, Y.I., Tamar, A., Harb, J., Abbeel, P., Mordatch, I.: Multi-agent actor-critic for mixed cooperative-competitive environments. In: Advances in Neural Information Processing Systems (NeurIPS). vol.~30 (2017)

\bibitem{RWARE}
Papoudakis, G., Christianos, F., Schäfer, L., Albrecht, S.V.: Benchmarking multi-agent deep reinforcement learning algorithms in cooperative tasks. In: NeurIPS Datasets and Benchmarks (2021)

\bibitem{rw11}
Rashid, T., Samvelyan, M., de~Witt, C.S., Farquhar, G., Foerster, J.N., Whiteson, S.: Qmix: Monotonic value function factorisation for deep multi-agent reinforcement learning. In: Thirty-Fifth International Conference on Machine Learning (ICML). pp. 4292--4301 (2018)

\bibitem{gae}
Schulman, J., Moritz, P., Levine, S., Jordan, M.I., Abbeel, P.: High-dimensional continuous control using generalized advantage estimation. In: 4th International Conference on Learning Representations (ICLR) (2016)

\bibitem{ppo}
Schulman, J., Wolski, F., Dhariwal, P., Radford, A., Klimov, O.: Proximal policy optimization algorithms. arXiv preprint arXiv:1707.06347  (2017)

\bibitem{it1}
Tang, H., Zhang, H., Shi, Z., Chen, X., Ding, W., Zhang, X.P.: Autonomous swarm robot coordination via mean-field control embedding multi-agent reinforcement learning. In: IEEE/RSJ International Conference on Intelligent Robots and Systems (IROS). pp. 8820--8826 (2023). \doi{10.1109/iros55552.2023.10341749}

\bibitem{rw12}
Wang, J., Ren, Z., Liu, T., Yu, Y., Zhang, C.: Qplex: Duplex dueling multi-agent q-learning. In: Ninth International Conference on Learning Representations (ICLR) (2021)

\bibitem{rw16}
Wang, T., Zeng, L., Dong, W., Yang, Q., Yu, Y., Zhang, C.: Context-aware sparse deep coordination graphs. In: Tenth International Conference on Learning Representations (ICLR) (2022)

\bibitem{rw10}
Yu, C., Velu, A., Vinitsky, E., Gao, J., Wang, Y., Bayen, A., Wu, Y.: The surprising effectiveness of ppo in cooperative multi-agent games. In: Advances in Neural Information Processing Systems (NeurIPS). vol.~35, pp. 24611--24624 (2022)

\bibitem{rw25}
Zhang, B., Bai, Y., Xu, Z., Li, D., Fan, G.: Efficient policy generation in multi-agent systems via hypergraph neural network. In: 29th International Conference on Neural Information Processing (ICONIP). pp. 219--230 (2022). \doi{10.1007/978-3-031-30108-7\_19}

\bibitem{it2}
Zhang, R., Zong, Q., Zhang, X., Dou, L., Tian, B.: Game of drones: Multi-uav pursuit-evasion game with online motion planning by deep reinforcement learning. IEEE Transactions on Neural Networks and Learning Systems  \textbf{34}(10),  7900--7909 (2022). \doi{10.1109/tnnls.2022.3146976}

\bibitem{rw21}
Zhu, S., Zhou, J., Chen, A., Bai, M., Chen, J., Xu, J.: Maexp: A generic platform for rl-based multi-agent exploration. In: IEEE International Conference on Robotics and Automation (ICRA). pp. 5155--5161 (2024). \doi{10.1109/ICRA57147.2024.10611573}

\bibitem{rw24}
Zhu, T., Shi, X., Xu, X., Gui, J., Cao, J.: Hypercomm: Hypergraph-based communication in multi-agent reinforcement learning. Neural Networks  \textbf{178},  106432 (2024). \doi{10.1016/J.NEUNET.2024.106432}

\end{thebibliography}

\end{document}